%% file: templateArxiv.tex
\documentclass{article}

\usepackage{PRIMEarxiv}

\usepackage[utf8]{inputenc} 
\usepackage[T1]{fontenc}    
\usepackage{hyperref}       
\usepackage{url}            
\usepackage{booktabs}       
\usepackage{amsfonts}       
\usepackage{nicefrac}       
\usepackage{microtype}      
\usepackage{lipsum}
\usepackage{fancyhdr}       
\usepackage{graphicx}       
\graphicspath{{media/}}     
\usepackage{amsmath}
\usepackage{enumitem}
\usepackage{bm}
\usepackage{xcolor}
\usepackage{color, colortbl}
\usepackage{graphics}

\usepackage{microtype}

\usepackage{inconsolata}
\usepackage{boxedminipage}
\usepackage{pifont}
\definecolor{shadecolor}{RGB}{237,237,237}
\newcommand{\mybox}[1]{\par\noindent\colorbox{shadecolor}
{\parbox{\dimexpr\textwidth-2\fboxsep\relax}{#1}}}

\begin{document}

\title{CohortGPT: An Enhanced GPT for Participant Recruitment in Clinical Study
}
\author{
  Zihan Guan \footnote[2]{}\\
  School of Computing \\
  University of Georgia \\
  \texttt{zihan.guan@uga.edu} \\
   \And
  Zihao Wu \footnote[2]{} \\
  School of Computing \\
  University of Computing \\
  \texttt{zw63397@uga.edu} \\
  \And
  Zhengliang Liu \\
  School of Computing \\
  University of Georgia \\
  \texttt{zl18864@uga.edu} \\
  \And
  Dufan Wu \\
  Department of Radiology \\
   Massachusetts General Hospital and Harvard Medical School \\
  \texttt{dwu6@mgh.harvard.edu} \\
  \And
  Hui Ren\\
  Department of Radiology \\
   Massachusetts General Hospital and Harvard Medical School \\
  \texttt{hren2@mgh.harvard.edu} \\
  \And
  Quanzheng Li\\
  Department of Radiology \\
   Massachusetts General Hospital and Harvard Medical School \\
  \texttt{li.quanzheng@mgh.harvard.edu} \\
  \And
  Xiang Li \\
  Department of Radiology \\
   Massachusetts General Hospital and Harvard Medical School \\
  \texttt{xli60@mgh.harvard.edu} \\
  \AND
  Ninghao Liu  \thanks{Corresponding Author} \\
  School of Computing \\
  University of Georgia \\
  \texttt{ninghao.liu@uga.edu} \\
}

\maketitle
\begin{abstract}
Participant recruitment based on unstructured medical texts such as clinical notes and radiology reports has been a challenging yet important task for the cohort establishment in clinical research. Recently, Large Language Models (LLMs) such as ChatGPT have achieved tremendous success in various downstream tasks thanks to their promising performance in language understanding, inference, and generation. It is then natural to test their feasibility in solving the cohort recruitment task, which involves the classification of a given paragraph of medical text into disease label(s). However, when applied to knowledge-intensive problem settings such as medical text classification, where the LLMs are expected to understand the decision made by human experts and accurately identify the implied disease labels, the LLMs show a mediocre performance. A possible explanation is that, by only using the medical text, the LLMs neglect to use the rich context of additional information that languages afford. To this end, we propose to use a knowledge graph as auxiliary information to guide the LLMs in making predictions. Moreover, to further boost the LLMs adapt to the problem setting, we apply a chain-of-thought (CoT) sample selection strategy enhanced by reinforcement learning, which selects a set of CoT samples given each individual medical report. Experimental results and various ablation studies show that our few-shot learning method achieves satisfactory performance compared with fine-tuning strategies and gains superb advantages when the available data is limited. The code and sample dataset of the proposed CohortGPT model is available at: \href{https://anonymous.4open.science/r/CohortGPT-4872/}{https://anonymous.4open.science/r/CohortGPT-4872/}
\end{abstract}



\section{Introduction}

Randomized Clinical Trials (RCTs) are a crucial component of evidence-based medicine for evaluating the efficacy of new biological agents, drugs, devices, or procedures in preventing or treating diseases~\cite{spieth2016randomized}. The completion of trials can be impeded by various obstacles, with participant recruitment often identified as a primary barrier~\cite{lai2019review} due to the potentially limited accessibility to the specific target group and fitting research recruitment into daily practice~\cite{foster2015barriers}, plus the difficulty in identifying individuals who meet all the inclusion and exclusion criteria outlined by the trial design, especially when the criteria items are not routinely recorded in the medical record~\cite{kaplan2013clinical}. While an increasing number of studies are utilizing structured electronic medical record (EMR) data for recruiting participants (i.e., EMR-enhanced recruitment)~\cite{cuggia2011comparing,kopcke2014employing}, the utilization of unstructured or semi-structured text data such as clinical notes and radiology reports to identify potential participants is still a challenging task due to the inherent complexity and variability of medical text used. Clinical notes and radiology reports often contain abbreviations, medical jargon, typographical errors, and inconsistent formatting, making it difficult to accurately and efficiently identify the information related to the enrollment criteria~\cite{hassanzadeh2020matching}. Additionally, the lack of standardized terminology and varying documentation styles make the process further complicated. 

In response to the above challenges, there have been increasing studies utilizing techniques in Natural Language Processing (NLP), especially text classification methods, to identify suitable enrollment participants~\cite{ismail2023role}. Text classification plays a pivotal role in NLP~\cite{aggarwal2012survey}, given its extensive applicability in real-world scenarios and congruity with various specialized domains, encompassing customer segmentation, recommendation systems, and outcome prediction. Additionally, it acts as an ideal standard for assessing the language comprehension capacity of a model. Eight of the nine tasks in the popular GLUE benchmark are classification tasks~\cite{wang2018glue}. Text classification in healthcare NLP has been extensively investigated to facilitate patient outcome prediction~\cite{goh2021prediction,lu2021natural}, computer-aided diagnosis~\cite{qomariyah2022nlp}, and hospital management~\cite{ong2010automated}.

In previous studies, participant recruitment through text classification was mainly centered around rule-based methods and machine-learning techniques. Despite its advantages of rapid inference and elimination of supervised training, the rule-based approach necessitates abundant data to extract statistical information, such as lexical frequency summation and class correspondence~\cite{wu2018semehr}. Additionally, the involvement of medical experts in the rule-making process is critical, as it mandates specialized knowledge to ensure the rules' alignment with anticipated classification outputs~\cite{harrer2019artificial}. With the advancement of deep learning, particularly the transformer module such as BERT~\cite{acheampong2021transformer}, more research has been performed using machine-learning approaches for participant recruitment~\cite{tian2021transformer,hassanzadeh2020matching,cai2021improving,widera2019machine,pakhomov2005prospective}. As BERT is pre-trained on large unlabeled datasets for improved capability in modeling language, it can be subsequently fine-tuned on labeled data to adapt to specific downstream tasks. With the increased availability of large-scale medical text data on the web, several BERT variants, such as BioBERT~\cite{lee2020biobert} and clinicalBERT~\cite{huang2019clinicalbert}, have emerged, pre-trained on publicly accessible medical text and clinical notes. However, employing pre-trained models for downstream tasks like participant enrollment necessitates a substantial quantity of labeled data for fine-tuning, which can be time-consuming and labor-intensive to obtain~\cite{ngiam2019big}. 

Recently, large Language Models (LLMs) such as ChatGPT and GPT-4 have achieved impressive language understanding and zero-shot in-context learning capabilities. Unlike BERT-based language models, LLMs feature a substantially larger model scale. They are pre-trained on extensive datasets using Reinforcement Learning from Human Feedback (RLHF), which aligns the models more closely with human expectations. ChatGPT and GPT-4 demonstrate human-like language understanding, reasoning, and generating capabilities, with impressive results on several open-domain NLP benchmarks without fine-tuning~\cite{fijavcko2023can,gilson2023does}. On the other hand, in highly specialized domains such as healthcare, their performance will be degraded since most LLMs were exclusively pre-trained on open-domain data, which lacks domain-specific vocabularies and knowledge~\cite{kitamura2023chatgpt}.

In response to these challenges, we developed an LLM-driven, radiology report-based participant recruitment framework called "CohortGPT." The proposed framework seamlessly integrates ChatGPT and GPT-4's impressive open-domain language understanding and reasoning abilities with specially designed prompting for medical domain tasks. Specifically, we embed medical and clinical knowledge into the ChatGPT model by utilizing a clinical-domain knowledge graph in the prompt design. Additionally, to optimize knowledge retrieval and reasoning capabilities, we employ Chain-of-Thought (CoT) prompting to guide the model to think step by step, thereby further bridging the domain knowledge gap. CohortGPT achieved competitive results compared with other deep learning-based methods using much less labeled data. The proposed framework can also be readily extended to other medical NLP tasks.

\section{Preliminary}
In this section, we present the preliminary problem statement and symbol notations used in this paper.
\subsection{Problem Statement}

Medical report classification could be formally defined as a multi-label classification problem as follows: Given a medical report $x_i \in \mathcal{D}$ composed of text diagnosis for the patients, we aim to learn a model which generates an answer that contains all the potential diseases implied by the report. For example, given the medical report [\textit{No acute disease. The heart is normal in size. The mediastinum is unremarkable. The lungs are clear.}], the model is expected to classify the report to the class of "\textit{Normal/No Disease}". Given the medical report [\textit{\underline{Mild cardiomegaly}. Normal pulmonary vascularity. No focal infiltrate, pneumothorax, or pleural effusion.}], the model is expected to classify the report as the class of "Cardiomegaly Disease".

\subsection{Symbol Notation}
Given a dataset $\mathcal{D}$, which could be further split into the training subset $\mathcal{D}_{train} \subset \mathcal{D}$ and the test subset $\mathcal{D}_{test} \subset \mathcal{D}$, we aim to utilize an LLM $f_{\theta}$ with fixed parameter $\theta$ to classify the reports in the test subset $\mathcal{D}_{test}$. The model $f_{\theta}$'s input is denoted as a sequence of texts $t = (t_1, t_2,...,t_n), t \in \mathcal{D}_{test}$.

\paragraph{Chain-of-Thought (CoT) prompting.} CoT prompting is proposed to address the cases where the input-output mapping is non-trivial. The key idea is that some examples of "questions" $q_i$ and "answers" $a_i$ are introduced to the input, detailing how the correct answers are deducted from the given information. Formally, a CoT prompting function is denoted as $p^{cot}(t) = (q_1, a_1, ..., q_m, a_m, t_1,t_2...,t_n)$. In this way, a query to the LLM is mapped to $f_{\theta}(p^{cot}(t))$.

\paragraph{Few-shot Learning.}
For the traditional few-shot learning, a few numbers of training samples $\{x_i|x_i \in \mathcal{D}_{train}, 0\leq i\leq k\}$ could be accessed. In this paper, following~\cite{brown2020language}, we denote the method as $k-$shot if $k$ samples are used as a part of the prompt.

\section{Methodology}
\begin{figure*}[!h]
    \centering
    \includegraphics[width=\textwidth]{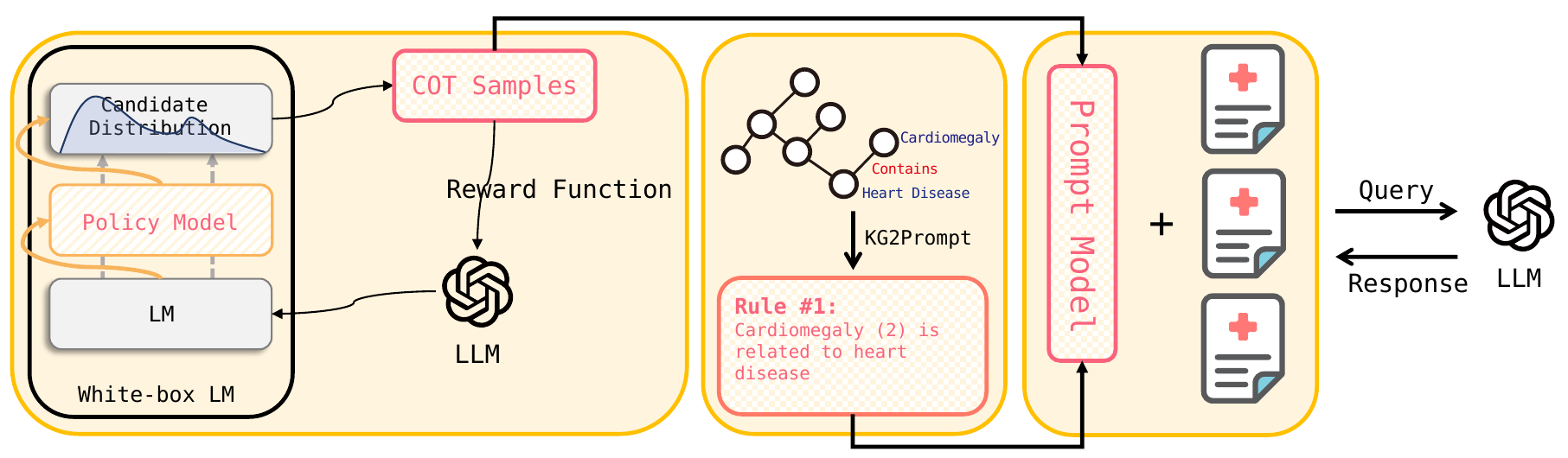}
    \caption{A policy model will be trained on a small number of training samples to dynamically select CoT samples from a CoT candidate pool. A knowledge graph containing the hierarchical information of the disease labels will be transformed into a series of executable rules. Then the dynamic CoT samples and the rules will be used to construct a prompt model. In the inference stage, the LLMs will be queried with the medical reports and the prompt model.}
    \label{fig:my_label}
\end{figure*}

\subsection{Static Prompting From Explicit Knowledge}
Despite the great advantages of understanding human languages, LLMs alone represent a limited coverage of knowledge. To bridge the gap, some alternate sources of information, such as knowledge graphs (KG) are usually used to enhance the reasoning ability of LLMs in the specific-domain downstream task~\cite{zhang2020radiology,he2021klmo,fei2021enriching,yasunaga2021qa}. In the following section, we give a detailed picture of how we embed the KG information into the input of LLMs.

\begin{figure}
    \centering
    \includegraphics[width=0.48\textwidth]{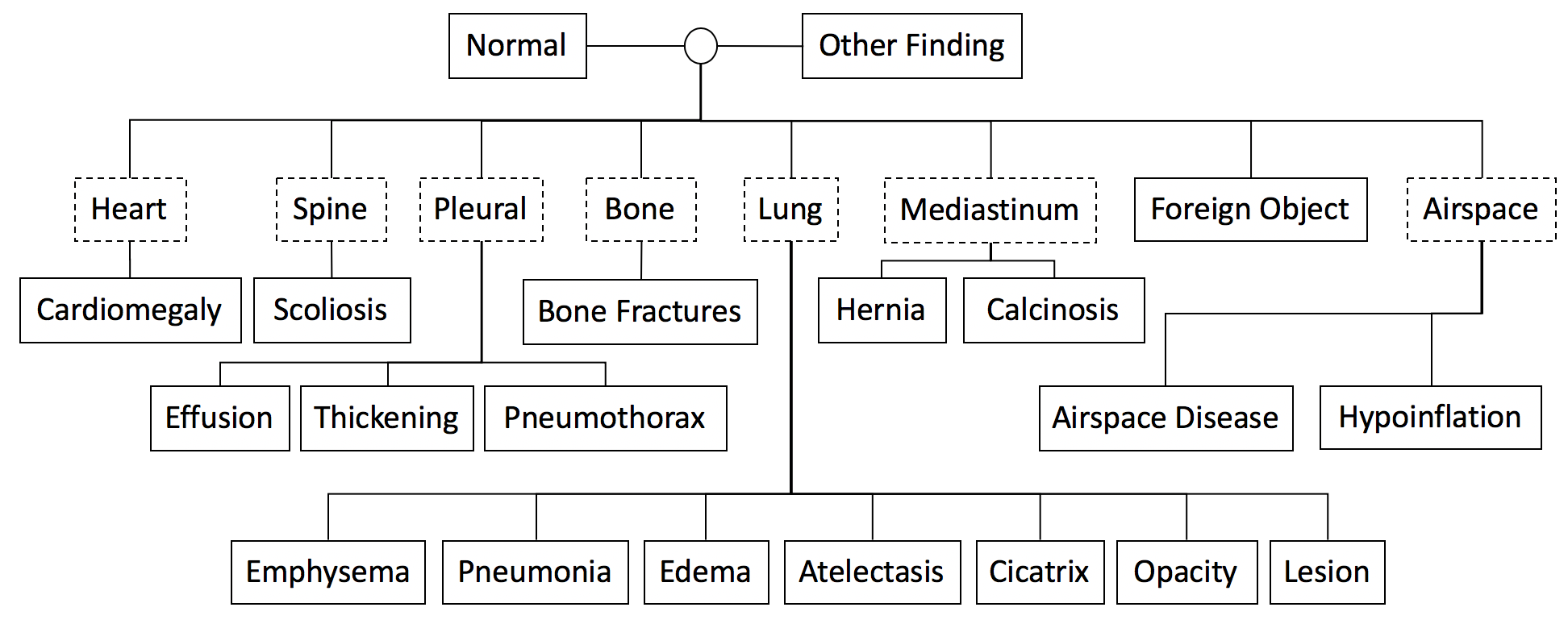}
    \caption{A knowledge graph was created by ~\cite{zhang2020radiology} to represent relationships between diseases, organs, or tissues. In this graph, disease labels are represented by nodes in solid boxes, corresponding organs or tissues are represented by nodes in dotted boxes, and the edges linking the nodes represent the relationships between disease keywords. Clusters are formed when disease labels are connected to the same tissue.}
    \label{fig:kg}
\end{figure}

\paragraph{Knowledge Graph}
Knowledge graphs are represented with many triplets of subject-predicate-object. In this paper, we consider a hierarchical knowledge graph as introduced in~\cite{zhang2020radiology} that covers the most common abnormalities or findings for the medical report diagnosis. Specifically, each node in the solid box denotes a disease label, each node in the dotted box denotes the corresponding organs or tissues, and each edge linking the nodes denotes the relationship between the disease keywords (e.g., "contains"). Disease labels that are linked to the same tissue or organ form a cluster. Figure~\ref{fig:kg} depicts an example of a knowledge graph constructed for the IU-RR dataset.

To embed the tree-structured knowledge into the LLMs, we propose several simple but effective prompt-based methods: KG-as-Tree, KG-as-Relation, and KG-as-Rule. KG-as-Tree aims to make the prompt maintain the tree-structured information in the KG and teach the LLM understand the hierarchical relationship among the various disease labels. Specifically, we use markdown-style symbols such as "\#", "\#\#" to denote different levels of the labels and point out that "the disease labels in the same level cannot be simultaneously chosen" at the end of the text. An excerpt of the prompt is as follows: "\# Heart \#\# Cardiomegaly, \# Spine \#\# Scoliosis ...". KG-as-Relation aims to transform the KG into a series of triplet relationships, e.g., "[Heart disease] [contains] [Cardiomegaly]; [Spine disease] [contains] [Scoliosis];...". Finally, KG-as-Rules decomposes the knowledge graph into a set of human-readable rules for LLMs. Specifically, we extract nine rules for each cluster of labels, e.g., for the mediastinum, we have '\emph{Rule \#7: hernia hiatal (8) and calcinosis(9) are both related to the Mediastinum disease}'. The detailed rules can be found in Appendix~\ref{appendix:rules}.







\subsection{Dynamic Prompting via Policy Gradient}
The in-context samples have shown great advantages in boosting the reasoning ability of LLMs~\cite{dong2022survey}, where the chain-of-thought (CoT) sample is one of the most effective types~\cite{wei2022chain}. Intuitively, the CoT samples guide the LLMs to learn the task-specific logical chains by providing examples of input and detailed output. Therefore, to further exploit the reasoning ability of LLMs in the medical report diagnosis task, we aim to incorporate CoT samples in the prompt.

The selection of CoT samples can be a random or a retrieved-based strategy. However, recent research has shown that the performance of the LLMs can be unstable with different selections or permutations of the CoT samples. Besides, the black-box setting for LLMs makes gradient information inaccessible to the model users. Therefore, it would be never non-trivial to consider the problem of \textit{CoT sample selection}. Inspired by~\cite{lu2022dynamic}, we adopt a policy-gradient-based strategy, where the selection of CoT samples is optimized with only feed-forward propagation.

Formally, given a medical report $x_i$, we aim to find $K$ CoT samples $\text{CoT}_i = \{c^1_i,c^2_i,...,c^K_i\}$ from the candidate pool $C_{cand}$ to construct the prompt $p_i$, where $p_i = [\text{CoT}_i, \kappa]$ is composed of the two critical elements: $\text{CoT}_i$ for CoT samples and $\kappa$ for the KG prompt. The choice of $c_i^k, k\in [1,K]$ follows a trainable policy neural network $\pi_{\theta}(\textbf{c}_i|x_i)$ as follows,
\begin{equation}
    c_i^k \overset{\mathrm{iid}}{\sim} \pi_{\theta}(\textbf{c}_i|x_i), s.t. \,\, c_i^k \in C_{cand},
\end{equation}
where $\pi_{\theta}(\textbf{c}_i|x_i) \in \mathbb{R}^{|C_{cand}|}$ denotes the sampling distribution over the set $C_{cand}$. 

Denoting the LLM as $f$, our goal is then to maximize the reward function as follows, which measures the performance of $f$ in the medical report classification task,
\begin{equation}
\begin{split}
        r_i &= \frac{1}{l} \sum_{l} [\lambda_1 \mathbb{I} (f(x_i, p_i)^l = y_i^l) \\
        & + \lambda_2 \mathbb{I} (f(x_i, p_i)^l \neq y_i^l)],
\end{split}
\end{equation}
where $\mathbb{I}$ is an indicator function that outputs -1 when the inner condition is false, and +1 when the inner condition is true, and $\lambda_1$ and $\lambda_2$ denote the coefficients for the correctly classified labels and incorrectly classified labels, respectively. Averaging across all the labels, we obtain the final reward given the input $x_i$ and the prompt $p_i$. Now our goal is to maximize the reward in the following problem,
\begin{equation}
    \max_{\theta} \,\, \mathbb{E}_{\text{CoT}_i \sim\pi_{\theta}(\mathbf{c}_i|x_i)}   r_i.
    \label{problem:reward_max}
\end{equation}
More details about the reward function is given in Appendix~\ref{appendix:reward}.
However, as mentioned previously, directly solving the Equation~\ref{problem:reward_max} is hard due to the inaccessible gradient information. By the virtue of policy gradient algorithm~\cite{sutton1998introduction} and the efficient implementation in PyTorch~\cite{schulman2016gradient}, the gradient could be estimated as
\begin{equation}
\begin{split}
        &\nabla_{\theta} \mathbb{E}_{\text{CoT}_i \sim \pi_{\theta}(\mathbf{c}_i|x_i)}   r_i \\
    \approx  & \frac{1}{M} \sum_{i=1}^{M} \nabla_{\theta} \log(\mathbb{P}(\text{CoT}_i|\pi_{\theta}(c_i|x_i))) \cdot r_i,
        \label{eqn:policy_gradient}
\end{split}
\end{equation}
where $M$ denotes the batch size and $\mathbb{P}$ denotes the probability.

Equation~\ref{eqn:policy_gradient} harnesses the logical similarity between the input sample $x_i$ and the CoT samples $\text{CoT}_i$ to boost the reward performance but omits their contextual similarity information. To this end, we introduce pre-trained language models~\cite{devlin2018bert} to capture the similarity between the input sample and the CoT samples. Specifically, we adopt token embedding from the BioGPT model~\cite{luo2022biogpt} as the status encoding. Then, we add an additional fully-connected layer to the top of the pre-trained language model to construct the policy neural network. Formally, the architecture can be written as follows,
\begin{equation}
    \begin{split}
        h(x_i) &= \mathbf{W} \cdot \text{BioGPT}(x_i) + \bm{b} , \\
        h({cand}_j) &= \mathbf{W} \cdot \text{BioGPT}({cand}_j) + \bm{b} ,\\
        \pi_{\theta}(\mathbf{c}_i^j|x_i) &= \frac{\exp(h(x_i) \cdot h({cand}_j))}{\sum_{c' \in C_{cand}} \exp(h(x_i) \cdot h(c'))} ,
    \end{split}
\end{equation}
where ${cand}_j \in C_{cand}$.
It is noted that the weights of the BioGPT model are fixed during the training stage. The only training parameters are $\mathbf{W}$ and $\bm{b}$.

\section{Experiment}
\subsection{Experimental Settings}
\paragraph{Dataset} To evaluate the effectiveness of the proposed method, we use two popular medical diagnosis datasets: IU-RR~\cite{demner2016preparing} and MIMIC-CXR~\cite{johnson2019mimic}.
\begin{itemize}[leftmargin=*]
    \item \textbf{IU-RR dataset} is a public dataset that contains 3955 radiology reports, each containing features such as 'findings', 'impression', 'MeSH', and so on. As in~\cite{zhang2020radiology}, we extracted the 20 most common diseases as the target labels, where each report is assigned one or more labels. Besides, we build the input text by concatenating contents in 'findings' and 'impression'. The detailed descriptions and examples of the dataset are given in Appendix~\ref{appendix:datasets}.
    \item \textbf{MIMIC-CXR} is a publicly available database that contains 227835 radiology reports, where each medical report contains features such as 'findings', 'impression" and so on. We split the whole dataset into training and testing sets as in the official setting. Then, we apply the same data processing pipeline as in~\cite{ma2023impressiongpt} to use CheXpert labeler~\cite{irvin2019chexpert} to assign pseudo labels for the medical reports and further filter out 1808 testing samples for evaluating the effectiveness of the proposed method.
\end{itemize}

\paragraph{Baseline Models and Methods}
Our baseline experiments are designed to contain the following two parts: 1) \textbf{Comparison with traditional fine-tuning strategies with the pre-trained models}. Specifically, We choose to use the pre-trained Bio-Bert~\cite{lee2020biobert} and BioGPT~\cite{luo2022biogpt} as the backbone and add additional layers for the multi-label classification task. The two models are then fine-tuned on the train split of the datasets and evaluated on the test split of the datasets. 2) \textbf{Comparison with other few-shot setting LLMs}. Specifically, we choose Alpaca~\cite{alpaca} and BloomZ~\cite{scao2022bloom}, as two strong LLM baselines to ChatGPT/GPT-4. The two LLMs are prompted with the proposed method in the paper and evaluated with the test split of the datasets.

\paragraph{Evaluation Metrics}
We adopt five popular metrics for the multi-label classification task as in~\cite{sorower2010literature}: Exact Match Ratio (MR), Precision (P), Recall (R), F1-Score (F), and Hamming Loss (HL).
\begin{itemize}[leftmargin=*]
    \item \textbf{Exact Match Ratio} is the portion of complete correct predictions, averaged across all instances,
    \item \textbf{Precision} is the proportion of predicted correct labels to the total number of actual labels, averaged over all instances,
    \item \textbf{Recall} is the proportion of predicted correct labels to the total number of predicted labels,
    \item \textbf{F1-Score} is the harmonic mean of precision and recall,
    \item \textbf{Hamming Loss} evaluates the average difference between predictions and ground truth.
\end{itemize}

\paragraph{Implementation Details} 
For all of the datasets, fine-tuned Bio-Bert and Bio-GPT models are trained on the train split and tested on the test split. The 5-shot-Alpaca, 5-shot-BloomZ, 5-shot-ChatGPT, and 5-shot-GPT-4, are prompted with the proposed method in the paper. Our default hyper-parameter for dynamic CoT sample selection are as follows: the size of the CoT Candidate pool is 25, the number of training samples is 160, the KG-to-prompt strategy is KG-as-Rule, and finally, the number of $k$-shot samples is 5. All these parameters are evaluated in the ablation studies.

\subsection{RQ1: Medical Report Classification Performance}

\begin{figure*}[!t]
    \centering
    \includegraphics[width=\textwidth]{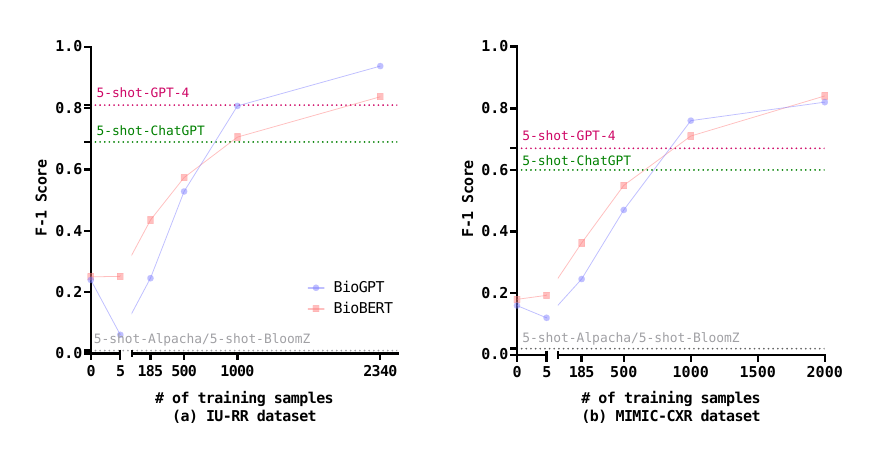}
    \caption{Effectiveness of the proposed method against the baseline methods.}
    \label{fig:GPT_BERT}
\end{figure*}

Figure~\ref{fig:GPT_BERT} presents the main results on the IU-RR dataset and MIMIC-CXR dataset, respectively. As shown, for the IU-RR dataset (Figure~\ref{fig:GPT_BERT} (a)), when only a limited number of training samples (e.g., 185) are accessible, the F1-Score performance of the proposed method integrated with ChatGPT (0.69) or GPT-4 (0.81) could outperform the traditional fine-tuning strategy (0.44 for BioBERT and 0.25 for BioGPT). However, the fine-tuning strategy eventually shows an advantage over the few-shot method when more training data samples are available. Similar results have also been observed in the experiments over the MIMIC-CXR dataset (Figure~\ref{fig:GPT_BERT} (b)), demonstrating that our method is advantageous in the few-shot setting.

\subsection{RQ2: Ablation Study}
\paragraph{Different KG-to-prompt Strategies}
Table~\ref{tab:kg_to_prompt_ablation} presents the comparison of different methods for embedding the KG into prompts. It is noted that the other parameters are fixed when adjusting the KG embedding method. As the table shows, the KG-as-Rule method exhibits the best performance across all metrics. This observation suggests that the LLMs such as ChatGPT are easier to handle command-style or rule-style inputs.
\input{table_kg_prompt_strategies}

\begin{figure*}[!t]
    \centering
    \includegraphics[width=\textwidth]{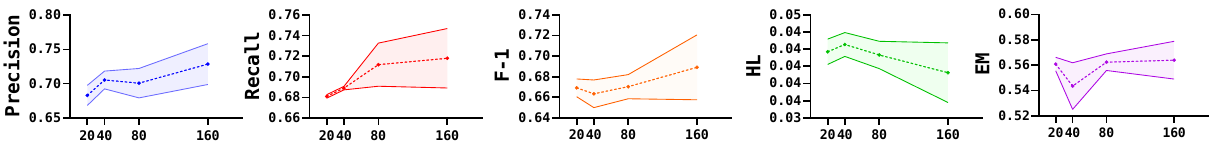}
    \caption{Impact on Number of Training Samples}
    \label{fig:training_samples_ablation}
\end{figure*}

\paragraph{Number of Training Samples}
Figure~\ref{fig:training_samples_ablation} presents the impact of the number of training samples on the performance of the proposed method, where the x-axis denotes different training samples and the y-axis denotes the metric evaluation. As the figure shows, with the increase of training samples, the precision, recall, F-1, EM all tend to be increasing, and HL tends to be decreasing. This also suggests that the performance of the proposed method tends to be enhanced with the increase of the training samples. A possible explanation for this observation is that, with more training samples, the policy model could be trained on a more generalized data distribution, leading to a better generalization ability on the testing set.

\begin{figure*}
    \centering
    \includegraphics[width=\textwidth]{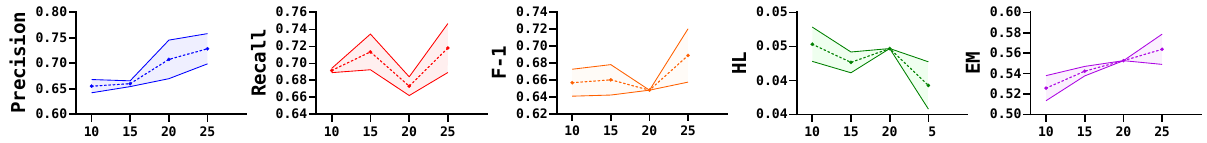}
    \caption{Impact on Number of Candidate Samples}
    \label{fig:candidation_samples}
\end{figure*}

\paragraph{Number of Candidate Samples}
Figure~\ref{fig:candidation_samples} presents the impact on the number of candidate samples. As the figure suggests, the performance becomes stronger with the increase in the number of candidate samples. This phenomenon could be interpreted as, with more samples in the candidate pool, the search space will be larger, leading the policy model more likely to escape the saddle points and achieve the optimum.

\begin{figure*}[!t]
    \centering
    \includegraphics[width=\textwidth]{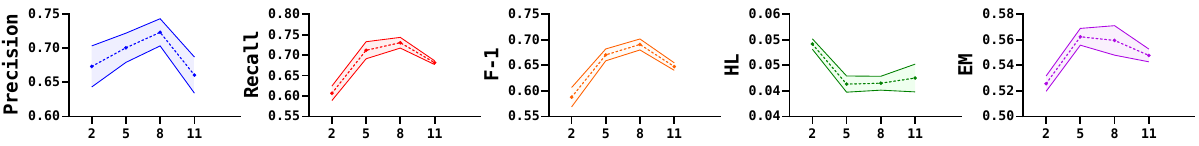}
    \caption{Impact on Number of $k$-shot samples}
    \label{fig:k_shot_ablation}
\end{figure*}

\paragraph{Number of $k$-shot samples}
Figure~\ref{fig:k_shot_ablation} presents the impact of the number of $k$-shot samples on the performance of the proposed method, where the x-axis denotes the k value, and the y-axis denotes the metric evaluation. As shown, the performance of the proposed method achieves its peak value when $k=5$ or $k=8$. After that, the performance goes down with a higher variance. We reckon that this is because excessive CoT samples lead the LLMs to be confusing as more chaotic information is introduced to the prompt.
\paragraph{Different CoT Selection Strategies}
Table~\ref{tab:CoT_selection_ablation} compares the proposed method with four other CoT selection strategies while keeping the other components (e.g., KG) in the prompt as fixed and only adjusting the CoT selection strategy. Random selection strategy randomly samples five CoT samples from the candidate pool. Manual selection uses the five fixed CoT samples whose text lengths are the longest, as they intuitively convey richer logical information. Most-similar selection strategy selects the 5 samples whose embedding is most similar to the given medical report. As the table suggests, the dynamic CoT selection strategy outperforms the other methods. Compared to the random selection and manual selection strategy, the great advantage stems from that the policy model could assign dynamic CoT samples given different reports. Compared to the most-similar selection strategy, our relative advantage comes from that the selected samples not only take the similarity into consideration but also the classification performance via maximizing the crafted reward function.

\input{table_selection_strategies}

\subsection{RQ3: Case Studies}
We present some examples of the selected CoT samples in the prompt and the corresponding answers generated by ChatGPT. As shown in Appendix~\ref{appendix:cot}, after prompting with these CoT samples, ChatGPT will mimically deduct the answers based on the given logic embeded in the CoT samples. Moreover, the selected CoT samples are not only "superficially" similar samples, but those with multiple reasoning steps as in the given test sample.

\section{Related Work}

\subsection{LLMs in healthcare}
The popular rise of transformer-based Large Language Models (LLMs) such as GPT-3 \cite{brown2020language} and GPT-4 \cite{openai2023gpt} has significantly transformed the landscape of natural language processing (NLP). Surpassing their precursors such as Recurrent Neural Networks \cite{tarwani2017survey,liu2022survey} and smaller pre-trained models (e.g., BERT \cite{devlin2018bert} or XLM \cite{lample2019cross}), these LLMs have expanded the horizons of performance across numerous tasks and demonstrate early signs of artificial general intelligence \cite{zhao2023brain,liu2023summary}. 

The objective of language models is to learn contextualized representations of the training text. For example, the word "dose" would have different meanings in a medical document versus in a culinary context. While previous models necessitate domain-specific pre-training \cite{gu2021domain} and fine-tuning \cite{devlin2018bert}, LLMs are inherently equipped to adapt to these contextual variations with minimal post-training adjustments, which enables LLMs to excel in few-shot or zero-learning \cite{brown2020language,liu2023summary}.

In the healthcare sector, the potential of LLMs is becoming increasingly evident. Extensive healthcare data \cite{liu2022survey,liu2023summary,arora2023promise}, encompassing clinical notes, patient records, and research articles, provides a fertile ground for LLMs to demonstrate their capabilities in classifying biomedical text \cite{chen2023evaluation}, data augmentation \cite{dai2023chataug}, de-identifying HIPAA-protected data \cite{liu2023deid}, summarizing radiology reports \cite{ma2023impressiongpt}, extracting clinical information \cite{agrawal2022large}, or depression and suicidality detection \cite{lamichhane2023evaluation}. 

Incorporating Reinforcement Learning from Human Feedback (RLHF) \cite{ouyang2022training} and instruction fine-tuning \cite{ouyang2022training} into Large Language Models (LLMs) significantly enhances their capacity to understand and align with individualized human values and communication nuances, which are pivotal in the healthcare, a domain that demands personal interactions, empathy and mutual understanding \cite{yu2022development}. This integration enables LLMs to better navigate the subtle complexities inherent in healthcare interactions and decision-making processes.

\subsection{Chain-of-thought reasoning with LLMs}
Chain-of-thought reasoning (CoT) is a problem-solving approach where complex problems are broken down into smaller, more manageable parts or steps \cite{wei2022chain,saparov2022language}. By addressing each part sequentially, the overall problem becomes easier to solve. This approach resembles how humans naturally tackle complicated problems, dividing them into simpler sub-problems and solving them one at a time. 

In the context of large language models, chain-of-thought reasoning aims to enhance the model's ability to generate more accurate and coherent responses by encouraging step-by-step reasoning processes \cite{wei2022chain}. For example, a zero-shot approach that simply requests the LLM to "think step by step" and concatenate its self-generated strategy to the subsequent prompt significant improves reasoning performance across a wide range of benchmarks \cite{kojima2022large}. When provided with a few more examples, LLMs can learn through in-context learning and achieve even better performance in reasoning \cite{wei2022chain}. 

More recent CoT implementations employs strategies from the broader machine learning domain to further unleash the potential of LLMs. Diao et al. \cite{diao2023active} proposed an active learning-inspired framework to improve large language model performance on reasoning tasks using an uncertainty-based annotation strategy. This approach involves calculating the uncertainty in the model's predictions, selecting the most uncertain questions for human annotation, and then using these annotated exemplars to enhance the model's reasoning abilities. The proposed Active-Prompt achieves state-of-the-art performance in arithmetic reasoning (e.g., 83.4 \% accuracy on the GSM8K dataset \cite{cobbe2021training}) and commonsense reasoning.

\section{Conclusion}

In this paper, we explore a new way to enhance LLM's inference ability in the medical domain by using a domain knowledge graph and an RL-enhanced CoT sample selection strategy. Experimental results and ablation studies show that the proposed framework could guide the LLMs to achieve satisfactory performance in a few-shot-learning setting compared with the fine-tuning strategy using much more labeled samples. While CohortGPT is based on ChatGPT and GPT-4, it can be implemented by any open-source LLMs, such as LLaMA, Vicuna, and Alpaca, which will greatly expand its feasibility by locally deployed. Furthermore, the text classification task investigated in this work can be readily explored in many other healthcare applications, including diagnosis, prognosis, and treatment optimization. 

\bibliographystyle{unsrt}  
\bibliography{references}  

\clearpage
\appendix
\section{More Details About Reward Function} \label{appendix:reward}
During training, we set the coefficients $\lambda_1 = 1$ and $\lambda_2 = -10$ as this combination is observed to yield satisfactory performance.

\section{More Details About Datasets} \label{appendix:datasets}
Figure~\ref{fig:appendix_dataset} presents an example of the medical report in the IU-RR dataset. The comparison, indication, findings, and Impression are the raw features of the medical report. As discussed in the previous section, we combine the texts in Findings and Impression to generate the final diagnosis report.
\input{appendix_dataset_examples}

\section{More Details About Rules} \label{appendix:rules}
Figure~\ref{fig:appendix_rules} presents the constructed rules by KG-as-Rule Method for the IU-RR dataset.
\input{appendix_rules_examples}


\section{Case Studies} \label{appendix:cot}
Figure~\ref{fig:selected_exp_promptpg} and Figure~\ref{fig:appendix_cot} present two examples of test samples, along with their selected CoT samples.
\input{appendix_cot_examples}

\end{document}

%% file: table_kg_prompt_strategies.tex
\begin{table*}[!t]
    \centering
    \resizebox{\textwidth}{!}{  
    \begin{tabular}{l|ccccc}
    \toprule
       \textbf{Methods}  &  \textbf{Exact Match Ratio} & \textbf{Precision} & \textbf{Recall} & \textbf{F1-score} & \textbf{Hamming Loss}$\downarrow$\\
       \midrule
        No KG & $0.54_{\pm 0.04}$ & $0.68_{\pm 0.02}$ & $0.71_{\pm 0.03}$ & $0.69_{\pm 0.03}$ & $0.05_{\pm 0.01}$\\
        \midrule
        KG-as-Tree  & $0.55_{\pm 0.02}$ & $0.69_{\pm 0.01}$ & $0.70_{\pm 0.02}$ & $0.69_{\pm 0.02}$ & $0.04_{\pm 0.01}$\\
        KG-as-Relation & $0.55_{\pm 0.02}$ & $0.70_{\pm 0.03}$ & $0.71_{\pm 0.03}$ & $0.70_{\pm 0.03}$ & $0.04_{\pm 0.01}$\\
        \textbf{KG-as-Rule} (default)  & $\textbf{0.56}_{\pm \textbf{0.01}}$ & $\textbf{0.73}_{\pm \textbf{0.02}}$ & $\textbf{0.72}_{\pm \textbf{0.02}}$ & $\textbf{0.69}_{\pm \textbf{0.03}}$ & $\textbf{0.04}_{\pm \textbf{0.01}}$\\
        \bottomrule
    \end{tabular}}
    \caption{Impact on Different Methods for transforming the knowledge graph information to prompt.}
    \label{tab:kg_to_prompt_ablation}
\end{table*}

%% file: table_selection_strategies.tex
\begin{table*}[!t]
    \centering
    \resizebox{\textwidth}{!}{  
    \begin{tabular}{l|ccccc}
    \toprule
       \textbf{Methods}  &  \textbf{Exact Match Ratio} & \textbf{Precision} & \textbf{Recall} & \textbf{F1-score} & \textbf{Hamming Loss $\downarrow$}\\
       \midrule
        Random Selection & $0.55_{\pm 0.02}$ & $0.64_{\pm 0.02}$ & $0.66_{\pm 0.03}$ & $0.62_{\pm 0.01}$ & $0.05_{\pm 0.01}$\\
        Manual Selection & $0.55_{\pm 0.02}$ & $0.65_{\pm 0.02}$ & $0.66_{\pm 0.04}$ & $0.62_{\pm 0.03}$ & $0.04_{\pm 0.01}$\\
        Most Similar & $0.56_{\pm \textbf{0.01}}$ & $0.65_{\pm 0.01}$ & $0.68_{\pm 0.03}$ & $0.64_{\pm 0.02}$ & $0.04_{\pm 0.01}$\\
        \textbf{Dynamic} & $\textbf{0.56}_{\pm \textbf{0.01}}$ & $\textbf{0.73}_{\pm \textbf{0.02}}$ & $\textbf{0.72}_{\pm \textbf{0.02}}$ & $\textbf{0.69}_{\pm \textbf{0.03}}$ & $\textbf{0.04}_{\pm \textbf{0.01}}$\\
        \bottomrule
    \end{tabular}}
    \caption{Impact on different CoT Selection Strategies}
    \label{tab:CoT_selection_ablation}
\end{table*}

%% file: appendix_dataset_examples.tex
\begin{figure*}[!t]
\centering
\fontsize{9.0pt}{\baselineskip}\selectfont
\begin{boxedminipage}{1.0\textwidth}
\begin{center}
    \mybox{
    \begin{minipage}[s][6cm]{0.98\textwidth}
        \vspace{-1mm}

         \textbf{comparison:} \\
         "Chest radiographs XXXX."
         
         \textbf{indication}\\
         "XXXX-year-old male, chest pain."

        \textbf{\color{red}{Findings}:} \\
        The cardiomediastinal silhouette is within normal limits for size and contour. The lungs are normally inflated without evidence of focal airspace disease, pleural effusion, or pneumothorax. Stable calcified granuloma within the right upper lung. No acute bone abnormality.. 

        \textbf{\color{red}{Impression:}} \\
        No acute cardiopulmonary process.\\
        \end{minipage}

        \begin{minipage}[s][2cm]{0.98\textwidth}
        \vspace{-1mm}
        \textbf{\color{blue}{Diagnosis Report:}} \\
        The cardiomediastinal silhouette is within normal limits for size and contour. The lungs are normally inflated without evidence of focal airspace disease, pleural effusion, or pneumothorax. Stable calcified granuloma within the right upper lung. No acute bone abnormality..No acute cardiopulmonary process.
        \end{minipage}

    }
\end{center}
\end{boxedminipage}
\caption{An example of the constructed diagnosis report by combining the texts in Findings and Impression of the raw dataset.}
\label{fig:appendix_dataset}
\end{figure*}

%% file: appendix_rules_examples.tex
\begin{figure*}[!t]
\centering
\fontsize{9.0pt}{\baselineskip}\selectfont
\begin{boxedminipage}{1.0\textwidth}
\begin{center}
    \mybox{
    \begin{minipage}[s][4.5cm]{0.98\textwidth}
        \vspace{-1mm}
         1. A report must not be classified into 'normal (1)' and disease labels 2-20 simultaneously! \\ 2. A report must not be classified into 'other findings(20)' and disease labels 1-19 simultaneously. \\ 3. Cardiomegaly (2) is related to heart disease. \\ 4. scoliosis / degenerative (3) is related to spine disease.\\ 5. fractures bone (4) is related to the bone disease. \\ 6. pleural effusion(5) thickening(6), and pneumothorax(7) are all related to the pleural disease\\7. hernia hiatal (8) and calcinosis(9) are both related to the Mediastinum disease. \\ 8. emphysema / pulmonary emphysema(10) pneumonia / infiltrate / consolidation(11) pulmonary edema(12) pulmonary atelectasis (13) cicatrix(14) opacity(15), and nodule / mass(16) are all related to lung disease.\\ 9. airspace disease(17), and hypoinflation / hyperdistention(18) are both related to airspace disease. 
        \end{minipage}
    }
\end{center}
\end{boxedminipage}
\caption{Constructed Rules by KG-as-Rule Method.}
\label{fig:appendix_rules}
\end{figure*}

%% file: appendix_cot_examples.tex
\begin{figure*}[!t]
\centering
\fontsize{9.0pt}{\baselineskip}\selectfont
\begin{boxedminipage}{1.0\textwidth}
\begin{center}
    \mybox{
    \begin{minipage}[s][6cm]{0.98\textwidth}
        \vspace{-1mm}
        \textbf{Q1:} \\
        Below is the medical report: [1. No acute cardiopulmonary abnormalities. 2. Emphysema and chronic bony abnormalities are unchanged from prior exams. .. The trachea is midline. The cardiomediastinal silhouette is normal. The superior thoracic spine is again noted, unchanged from prior. Lucent pulmonary parenchyma is consistent appearance with emphysema and appears unchanged from prior examinations. No evidence of pneumothorax. No focal airspace disease or pleural effusion. Vague density in the medial right lung apex most XXXX representing overlying shadows of bony structures, which is stable.] \\

        \textbf{A1:} \\
        The report mentions that 'Worsening bibasilar subpleural interstitial opacities', suggesting opacity(15); The report mentions 'Lung volumes are low', suggesting hypoinflation / hyperdistention(18); The report mentions 'There calcifications of the thoracic aorta.', suggesting calcinosis(9). Therefore, the output is [the disease indices are: (9, 15, 18)]\\
        \end{minipage}

    \begin{minipage}[s][5cm]{0.98\textwidth}
        \vspace{-1mm}
        \textbf{Q2:} \\
        Below is the medical report: [No acute cardiopulmonary abnormality... Cardiomediastinal silhouette is within normal limits for size, with redemonstration of tortuous and atherosclerotic calcified thoracic aorta. No focal consolidation, effusion, or pneumothorax identified. Eventration of the right hemidiaphragm is stable compared to prior examination. Multilevel degenerative disc disease and thoracolumbar spine again noted without acute osseous abnormality.]' \\

        \textbf{A2:} \\
        The report mentions that 'Multilevel degenerative disc disease and thoracolumbar spine', suggesting the scoliosis / degenerative. Therefore, the output is [the disease indices are: (3)]
        \end{minipage}



        \begin{minipage}[s][8.5cm]{0.98\textwidth}
        \vspace{-1mm}
        \textbf{Test Sample:} \\
        1. XXXX right apical pneumothorax measuring approximately 5 mm in thickness. 2. Multiple right-sided rib fractures involving at XXXX the right anterior 5th through 9th ribs with mild displacement. 3. Mild right basilar airspace disease, atelectasis versus contusion. .. The cardiomediastinal silhouette and pulmonary vasculature are within normal limits in size. There is a thin right apical pneumothorax measuring approximately 5 mm in thickness. There is extensive subcutaneous emphysema in the right chest wall and neck. There are fractures of the right anterior 5th through 9th anterior ribs with mild displacement. Additional fractures cannot entirely be excluded. There is mild streaky airspace disease in the right lung base. Left lung is clear. There is a small hiatal hernia. There is an intrathecal catheter terminating in the lower thoracic spine. \\

        \textbf{Answer:} \\
        The report mentions that 'XXXX right apical pneumothorax measuring approximately 5 mm in thickness', suggesting pneumothorax(7); the report mentions that 'Multiple right-sided rib fractures involving at XXXX the right anterior 5th through 9th ribs with mild displacement', suggesting fractures bone(4); the report mentions that 'Mild right basilar airspace disease, atelectasis versus contusion', suggesting airspace disease(17) and pulmonary atelectasis(13); the report mentions that 'There is extensive subcutaneous emphysema in the right chest wall and neck', suggesting emphysema / pulmonary emphysema(10); the report mentions that 'There is mild streaky airspace disease in the right lung base', suggesting airspace disease(17). According to the rule 6, 'pleural effusion(5), thickening(6), and pneumothorax(7) are all related to pleural disease'. Therefore, the output is [the disease indices are: (4, 7, 10, 13, 17)]
        \end{minipage}

    }
\end{center}
\end{boxedminipage}
\caption{An example of a test sample in the IU-RR dataset and the two corresponding CoT samples selected by the dynamic algorithm.}
\label{fig:selected_exp_promptpg}
\end{figure*}

\begin{figure*}[ht!]
\centering
\fontsize{9.0pt}{\baselineskip}\selectfont
\begin{boxedminipage}{1.0\textwidth}
\begin{center}
    \mybox{
    \begin{minipage}[s][5cm]{0.98\textwidth}
        \vspace{-1mm}
        \textbf{Q1:} \\
        Below is the medical report: [1. No acute cardiopulmonary findings.. The heart size and mediastinal contours appear within normal limits. Atherosclerotic calcification of the aorta. No focal airspace consolidation, pleural effusions or pneumothorax. Questionable thin-walled cavitary lesion in the right lower lobe, only seen on the AP view and may represent artifact. No acute bony abnormalities.] \\

        \textbf{A1:} \\
        The report mentions 'Atherosclerotic calcification of the aorta', but this disease does not belong to any of the potential labels from 2 to 19. Besides, according to the rule 2, 'A report must not be classified into 'other findings (20)' and disease labels 1-19 simultaneously'. Therefore, the output is [the disease indices are: (20)]\\
        \end{minipage}

    \begin{minipage}[s][5cm]{0.98\textwidth}
        \vspace{-1mm}
        \textbf{Q2:} \\
        Below is the medical report: [No acute cardiopulmonary abnormality... Cardiomediastinal silhouette is within normal limits for size, with redemonstration of tortuous and atherosclerotic calcified thoracic aorta. No focal consolidation, effusion, or pneumothorax identified. Eventration of the right hemidiaphragm is stable compared to prior examination. Multilevel degenerative disc disease and thoracolumbar spine again noted without acute osseous abnormality.] \\

        \textbf{A2:} \\
        The report mentions that 'Multilevel degenerative disc disease and thoracolumbar spine', suggesting the scoliosis / degenerative. Therefore, the output is [the disease indices are: (3)]
        \end{minipage}



        \begin{minipage}[s][4cm]{0.98\textwidth}
        \vspace{-1mm}
        \textbf{Test Sample:} \\
        No acute cardiopulmonary findings.. The cardiomediastinal silhouette and pulmonary vasculature are within normal limits in size. The lungs are clear of focal airspace disease, pneumothorax, or pleural effusion. There are no acute bony findings. \\

        \textbf{Answer:} \\
        The report mentions 'No acute cardiopulmonary findings', indicating a normal report (1). However, the report cannot be classified into 'normal (1)' and disease labels 2-20 simultaneously according to Rule 1. Therefore, the output is [the disease indices are: (1)].
        \end{minipage}

    }
\end{center}
\end{boxedminipage}
\caption{An example of a test sample in the IU-RR dataset and the two corresponding CoT samples selected by the dynamic algorithm.}
\label{fig:appendix_cot}
\end{figure*}